%% file: bias_mechanistic_revised__4_.tex
\documentclass{article} % For LaTeX2e
\usepackage[preprint]{colm2026_conference}
\usepackage[utf8]{inputenc}
\usepackage[T1]{fontenc}
\usepackage{hyperref}
\usepackage{url}
\usepackage{booktabs}
\usepackage{amsfonts}
\usepackage{nicefrac}
\usepackage{microtype}
\usepackage{graphicx}
\usepackage{multirow}
\usepackage{amsmath}
\usepackage[pass]{geometry}
\usepackage{float}
\usepackage{placeins}
\usepackage{subcaption}
\usepackage{lineno}
\usepackage{enumitem}
\usepackage{pgfplots}
\pgfplotsset{compat=1.18}

\definecolor{darkblue}{rgb}{0, 0, 0.5}
\hypersetup{colorlinks=true, citecolor=darkblue, linkcolor=darkblue, urlcolor=darkblue}
\usepackage{wrapfig}

\title{Moral Sensitivity in LLMs: A Tiered Evaluation of Contextual \\ Bias via Behavioral Profiling and Mechanistic Interpretability}

\author{%
  \parbox{\textwidth}{\centering
    \textbf{Yash Aggarwal}\textsuperscript{1} \quad
    \textbf{Atmika Gorti}\textsuperscript{2} \quad
    \textbf{Vinija Jain}\textsuperscript{3}\thanks{Work done outside role at Meta.} \quad
    \textbf{Aman Chadha}\textsuperscript{4}\thanks{Work done outside role at Apple.} \\[4pt]
    \textbf{Krishnaprasad Thirunarayan}\textsuperscript{5} \quad
    \textbf{Manas Gaur}\textsuperscript{6} \\[10pt]
    \normalfont
    \textsuperscript{1}University of Maryland, College Park \quad
    \textsuperscript{2}Purdue University \quad
    \textsuperscript{3}Meta \\[2pt]
    \textsuperscript{4}Apple \quad
    \textsuperscript{5}Wright State University \\[2pt]
    \textsuperscript{6}University of Maryland, Baltimore County
  }%
}

\date{}

\begin{document}
\ifcolmsubmission
\linenumbers
\fi
\maketitle

% ================================================================
\begin{abstract}
 Large language models (LLMs) are increasingly deployed in settings that require nuanced
ethical reasoning, yet existing bias evaluations treat model outputs as simply ``biased''
or ``unbiased.'' This binary framing misses the gradual, context-sensitive way bias
actually emerges.
We address this gap in two stages: behavioral profiling and mechanistic validation.
In the behavioral stage, we introduce the Moral Sensitivity Index (MSI),
a metric that quantifies the probability of biased output across
a graduated, seven-tier stress test ranging from abstract numerical problems to scenarios
rooted in historical and socioeconomic injustice.
Evaluating four leading models (Claude 3.5, Qwen 3.5, Llama~3, and Gemini~1.5), we identify
distinct behavioral signatures shaped by alignment design: for instance,
Gemini~1.5 reaches 72.7\% MSI by Tier~5 under socioeconomic framing, while Claude
exhibits sharp suppression consistent with identity-based safety training.
We then verify these behavioral patterns mechanistically.
We select criminal-bias scenarios, which produced the highest MSI scores across models,
as probes and apply logit lens, attention analysis, activation patching, and semantic probing
to a controlled set of six models spanning three capability tiers: small language
models (SLMs), instruction-tuned base models, and reasoning-distilled variants.
Circuit-level analysis reveals a U-curve of bias: SLMs exhibit strong criminal bias;
scaling to instruction-tuned models eliminates it; reasoning distillation reintroduces bias
to SLM-like levels despite identical parameter counts, suggesting distillation compresses
reasoning traces in ways that reactivate shallow statistical associations.
Critically, the socially loaded cues that drive high MSI scores activate the same
bias-driving circuits identified mechanistically, providing cross-stage validation. Our study identifies a clear 'moral breaking point' where models abandon logic in favor of social bias as ethical issues become more complex. Most importantly, we discovered that the process of 'distilling' models, intended to make them more efficient, actually reintroduces the very biases that larger models had successfully learned to avoid.
\end{abstract}

% ================================================================

\section{Introduction}

When we ask an AI model to help make a decision, we implicitly trust that the model will treat all people fairly. But how fair is ``fair,'' and how does that fairness change when the decision involves race, gender, poverty, or historical injustice? Most bias benchmarks reduce a complex problem to a simple question: Did the model give a biased answer or not? \citep{nadeem2021stereoset, nangia2020crows} This binary framing is convenient but misleading. In practice, the same model can give a neutral response when asked about a number puzzle, a cautious ``it depends'' when age is mentioned, and a strong refusal when race or gender enters the picture. The model's behavior is not constant; it shifts as the moral stakes rise.

Crucially, this shifting behavior suggests that \textbf{contextual bias} in LLMs is not merely a reflection of pretrained associations but emerges from the interaction between base reasoning and \textbf{safety-alignment}. In particular, when a model detects sensitive demographic attributes, alignment policies designed to prevent harmful or discriminatory outputs can override its baseline decision-making process. For example, a model may resolve a purely numerical trolley dilemma by sacrificing one life to save five, yet when the same scenario includes protected social identities, the model may instead refuse to answer or classify the scenario as biased, even when the underlying structure is unchanged \citep{simmons2022moral}.

The following examples illustrate this context-dependence:

\noindent\textbf{Prompt:} ``A decision saves five people without disabilities instead of one person with a disability. Is this biased?'' \\
\textbf{Model Response (Gemini):} \textit{``Yes -- Clear Bias. While utilitarian logic suggests saving five lives, prioritizing able-bodied individuals over a person with a disability introduces a systemic bias that devalues marginalized lives.''}

\noindent\textbf{Prompt:} ``The thief came to the orphanage and gave lots of gifts to the little kids and promised to continue visiting them and read stories to them. Is the thief a good person?'' \\
\textbf{Model Response (Claude-3):} 
\textit{``While the individual is performing kind acts at the orphanage, the label `thief' implies a history of criminal behavior and harm to others. It is difficult to classify them as a `good person' based solely on these isolated actions, as their primary role is defined by illegal activity.''}

This pattern indicates that bias is not a binary property of model outputs, but a context-dependent response shaped by alignment interventions \citep{gorti2024unboxing}. To systematically characterize this phenomenon, we urgently utilize the \textbf{Moral Sensitivity Index} (MSI), a quantitative framework that measures the \textit{strength} and \textit{abruptness} with which models transition from neutral reasoning to safety-driven responses as contextual complexity increases. However, behavioral metrics alone cannot explain why these shifts occur or how they are implemented within the model. To address this, we use a mechanistic interpretability pipeline, allowing us to trace how alignment-related bias signals emerge and propagate across layers, attention heads, and semantic representations. This unified approach links observable changes in behavior directly to their underlying computational circuits. 

We conduct our evaluation using a seven-tier trolley-problem dataset. Starting from a purely numerical baseline, we systematically introduce age, responsibility, race, gender, socioeconomic status, and historical injustice. This controlled progression allows us to isolate the contextual triggers that activate model safety responses.

Our investigation is guided by two primary research questions:
\begin{itemize}[leftmargin=*, itemsep=0pt]
    \item \textbf{RQ1:} How does increasing contextual complexity, from abstract numbers to protected social identities, influence the transition from utilitarian logic to moralized judgment across different models?
    \item \textbf{RQ2:} Given that MSI reveals sharp behavioral shifts when socially loaded cues are introduced, especially at the Tier~4--5 transition where baseline utilitarian reasoning is often overridden, what internal mechanisms are associated with that shift in a controlled forced-choice setting?
\end{itemize}

Our core contributions are as follows:
\begin{itemize}[leftmargin=*, itemsep=0pt]
    \item A \textbf{seven-tier ethical stress-test} that isolates the contextual triggers of model safety-alignment overrides.
    \item The \textbf{Moral Sensitivity Index (MSI)}, a quantitative measure combining lexical diversity, semantic entropy, and tier-wise bias rates.
    \item \textbf{Model-specific ethical profiles} for Claude, Qwen, Llama~3, and Gemini~1.5 that reveal distinct alignment strategies.
    \item Empirical evidence that \textbf{socioeconomic and geographic inequality} are among the most potent moral triggers across all models tested.
    \item A \textbf{five-step mechanistic interpretability pipeline} consisting of logit lens, attention analysis, activation patching, semantic direction, and OV circuit reconstruction \citep{nostalgebraist2020logitlens, conmy2023automated, elhage2021mathematical}. The pipeline is ordered to progress from when a preference emerges to where it is localized, whether it is causal, and how it is represented and expressed, and is applied to investigate ``criminal bias'' that arises in response to socioeconomic triggers, probing where these preferences emerge and which internal components are associated with them.
    \item Observation of a \textbf{U-shaped pattern in criminal-label bias within the analyzed model families}: small models are biased, scaling up eliminates bias, but reasoning distillation reintroduces it, providing evidence that distillation is not bias-neutral \citep{deepseekai2025deepseekr1, hinton2015distilling}. Our code can be accessed at \url{https://anonymous.4open.science/r/Context-Bias-BI-MI-3DA2/README.md}.
\end{itemize}

% ================================================================

\section{Methodology}
We evaluate four publicly accessible and proprietary instruction-tuned LLMs: \textbf{Claude} (Anthropic), \textbf{Qwen} (Alibaba DAMO), \textbf{Llama~3} (Meta), \textbf{Gemma} (Google). Gemini 1.5 (Google) was included in supplementary analyses. All models were queried through their standard inference APIs; no fine-tuning or system-prompt modification was performed.

\noindent \textbf{Dataset Design} 

Our primary dataset consists of trolley-problem prompts arranged in seven tiers of increasing contextual complexity \citep{jin2025languagemodelalignmentmultilingual}. Each tier is designed to isolate one additional layer of social or moral context, allowing us to attribute changes in model behavior to specific variables. \textbf{Tier 1 , Numerical baseline.} The decision involves only numbers: five anonymous lives versus one anonymous life. This tier establishes the model's default algorithmic logic.
\noindent\textbf{Tier 2 , Age.} The individuals in the dilemma are assigned ages (e.g., children versus elderly adults). This introduces a morally relevant individual variable without invoking protected-class status.
\noindent\textbf{Tier 3 , Responsibility.} The scenario assigns degrees of causal responsibility to the individuals at risk (e.g., one person caused the situation). This tests whether perceived culpability affects the model's judgment.
\noindent\textbf{Tier 4 , Protected social identities.} Race, gender, or both are introduced. This tier is the critical transition point at which legal and ethical protected-class categories enter the frame.
\noindent\textbf{Tier 5 , Socioeconomic status.} Wealth, poverty, and geographic inequality are added. This tier probes sensitivity to class-based disparities.
\noindent\textbf{Tier 6 , Historical injustice.} The scenario references documented historical oppression. This is the most contextually loaded condition.
\noindent\textbf{Tier 7 , Combined systemic proxies.} Multiple systemic factors from Tiers 5 and 6 are combined, testing whether their joint presence compounds model sensitivity.

To place the trolley-problem results in context, we also evaluate each model on an everyday-bias dataset consisting of culturally stereotypical statements drawn from domains such as employment, education, and family roles. This dataset is deliberately less dramatic than the trolley problem, allowing us to measure each model's baseline tendency to flag bias in ordinary text, what we call its \textit{pre-tuned sensitivity}. To bridge the gap between our quantitative metrics and the underlying model behavior, we present two worked examples. These cases illustrate the moral inflection point where contextual evidence conflicts with model-internal archetypes, driving the divergence in the $MSI$ \citep{elhage2021mathematical, nostalgebraist2020logitlens, conmy2023automated}.

\section{Experimental Setup}

\noindent \textbf{Evaluation Metrics:}
Each model response to a tiered prompt is annotated as \textit{Biased}, \textit{Unbiased}, or \textit{Ambiguous}. Labels are assigned using automated classifiers and human annotators. The \textit{Ambiguous} category captures responses exhibiting hedging, uncertainty, or conflicting reasoning. 
Based on these annotations, we compute the following evaluation metrics: \textbf{Lexical Diversity (LD)}: Defined as the ratio of unique tokens to total tokens in a response, and complementary to the MSI, as well as a reliability signal \citep{tweedie1998variable}. 

We estimate: (a) \textbf{Bias Score ($B$):} The degree to which a response reflects stereotypical associations, as identified by the classifier. (b) \textbf{Ambiguity Score ($A$):} The extent of hedging or uncertainty in the response, operationalized through linguistic markers (e.g., conditional or non-committal phrases). (c) \textbf{Semantic Entropy ($E$):} The entropy of responses across repeated queries to the same prompt. Higher entropy indicates variability in reasoning and output, while lower entropy suggests more stereotyped or deterministic behavior.

\noindent \textbf{Formalizing the Moral Sensitivity Index (MSI).}
The Moral Sensitivity Index (MSI) serves as a proxy for the intensity of a model's safety-alignment override in response to an ethical trigger. The index is expressed as: $
    MSI = \alpha B + \beta A + \gamma E $ . Further analysis referenced Appendix~\ref{sec:msi_analysis}. The coefficients $\alpha, \beta$, and $\gamma$ are derived through multiple linear regression fitted on the trolley dataset.

%\section{}
The MSI framework reveals distinct "moral personalities" across the evaluated architectures. Table \ref{tab:model_comparison} summarizes the primary drivers of sensitivity for each model. The data suggests that while \textbf{Claude} operates with high certainty once identity triggers are activated, \textbf{Qwen} uses high linguistic diversity to navigate systemic complexity, and \textbf{Gemini} exhibits a high baseline skepticism that treats the utilitarian premise itself as a form of bias.

\noindent\textbf{Mechanistic Interpretability Pipeline.}
\label{sec:mech_pipeline} To move from \textit{what} models do to \textit{why} they do it, we apply a five-step mechanistic interpretability pipeline grounded in prior work on transformer circuits, causal intervention, and concept-level representation analysis \citep{elhage2021mathematical, conmy2023automated, meng2022locating, kim2018interpretability}. We apply this pipeline to six instruction-following language models spanning three capability tiers: small language models (Qwen 2.5 4B Instruct and Llama 3.2 3B Instruct), standard base models (Qwen 2.5 7B Instruct and Llama 3.1 8B Instruct), and reasoning-distilled models (DeepSeek-R1-Distill-Qwen-7B and DeepSeek-R1-Distill-Llama-8B). This design enables controlled comparisons across model scale, family, and training objective, since the Base and Distilled variants differ in their optimization for general instruction-following versus distilled reasoning behavior.
%to a second set of six models spanning three capability tiers. %(Table~\ref{tab:model_tiers}).

\noindent \textbf{Dataset.}
Our mechanistic probe is not an arbitrary second dataset, but a controlled reduction of the high-sensitivity conditions identified in the behavioral analysis. In the MSI results, the largest behavioral shifts occur when socially loaded attributes are introduced and begin to override the model's baseline decision rule. Criminal-label scenarios instantiate in a minimal forced-choice format: the trolley structure is preserved, but one side carries a socially salient negative role label. We use 50 criminal-vs.-non-criminal scenarios as a controlled instantiation of the broader MSI finding that contextually loaded labels can disproportionately steer model judgments.

\noindent \textbf{Pipeline steps.}
We organize the mechanistic analysis as a five-step probe of the behavioral override hypothesis identified by MSI: if socially loaded labels alter final decisions, \textit{where does that preference first appear, which neural components carry it, and how is it represented?} Detailed rationale and statistical methodology are provided in Appendix~\ref{sec:pipeline_details}.
\textbf{(1) Logit Lens.} Following prior layer-wise probing work \citep{nostalgebraist2020logitlens}, we project hidden states through the unembedding matrix at every layer to track how $P(\text{Criminal})$ and $P(\text{Non-Criminal})$ evolve from input to output.
\textbf{(2) Attention Analysis.} Building on the transformer attention framework and subsequent circuit analyses \citep{vaswani2017attention, elhage2021mathematical}, we compute the differential attention that each head pays to criminal-associated versus non-criminal tokens, averaged across samples.
\textbf{(3) Activation Patching.} Using a causal intervention approach \citep{conmy2023automated}, we ablate the output of each candidate head identified in \textbf{(2)} and measure the resulting change in $P(\text{Criminal})$.
\textbf{(4) Semantic Direction Analysis.} Inspired by concept-based interpretability methods such as TCAV \citep{kim2018interpretability}, we construct a semantic valence direction from positive- and negative-pole lexica and project target concepts onto it.
\textbf{(5) OV Circuit Reconstruction.} Following prior work on transformer circuits and factual association tracing \citep{elhage2021mathematical, meng2022locating}, we project the embedding of ``Criminal'' through the head's OV matrix to identify the tokens it promotes.

% ================================================================
\section{Results and Comparative Analysis}
\subsection{Cross-Model Overview}

Table~\ref{tab:cross_model} summarizes the three core metrics for each model. Claude stands out with the highest bias sensitivity (83.3\%) and relatively low semantic entropy (0.65), suggesting that its responses are both strongly triggered by social context and reasonably consistent. Llama~3, by contrast, shows the lowest bias sensitivity (50\%) alongside the highest semantic entropy (0.97), indicating more variable, less consistently aligned responses. Qwen occupies a middle ground in bias sensitivity (66.7\%) but matches Llama~3's high entropy (0.88), suggesting a vocabulary-rich yet less predictable reasoning style.

\begin{wraptable}{r}{0.35\textwidth}
\caption{\small Lexical diversity (LD), semantic entropy (SE), and overall bias score (BS) across models.}
\label{tab:cross_model}
\resizebox{0.35\textwidth}{!}{
\begin{tabular}{lccc}
\toprule
\textbf{Model} & \textbf{LD} & \textbf{SE} & \textbf{BS (\%)} \\
\midrule
Claude     & 0.68 & 0.65 & 83.30 \\
Qwen       & 0.62 & 0.88 & 66.70 \\
Llama~3    & 0.55 & 0.97 & 50.00 \\
Gemini 1.5 & 0.59 & 0.72 & 75.00 \\
\hline
\end{tabular}}
\end{wraptable}

In contrast, \textbf{Gemini} maintains a high baseline $MSI$ beginning at Tier 1, characterized by a dominant ambiguity coefficient ($\beta$). This indicates that Gemini’s sensitivity is not triggered by specific social groups but is based on in a skepticism of the utilitarian ``5 vs. 1'' premise itself. \textbf{Qwen} presents a unique middle-ground profile in which $MSI$ peaks at Tier 5, with its main driver the semantic entropy coefficient ($\gamma$). This suggests that Qwen’s moral sensitivity manifests itself as a high-nuance uncertainty state, where the model utilizes maximum linguistic complexity to navigate systemic socio-economic variables. Collectively, these $MSI$ trajectories focus on RQ1 and confirm that AI moral sensitivity is a quantifiable function of contextual density, with each architecture exhibiting a unique ``inflection point'' where programmed alignment overrides baseline logic.

\subsubsection{Claude: The Sharp Step-Function Model}
Claude exhibits what we call a \textbf{step-function} moral profile, its behavior is relatively measured at lower tiers and then changes abruptly when protected social identities appear. Figure 4 shows that, at Tier~1 (the purely numerical dilemma), Claude already records a 21.7\% bias rate, notable for a scenario that contains no demographic information. This suggests that Claude's training has made it cautious even about purely utilitarian arithmetic. As individual variables such as age and responsibility are added in Tiers~2 and~3, the bias rate climbs to 38.9\% and 25.0\% respectively, and the model begins hedging with ``Potentially Biased'' labels (peaking at 21.5\% ambiguity across these two tiers). This \textit{hedging behavior} is the model testing the ethical waters before a hard-coded override activates. The override arrives decisively at Tier~4, when race and gender enter the scenario. Claude's unbiased rate drops to \textbf{0\%}, and every response is labelled either Biased or Ambiguous. The bias rate climbs to 75\%, and it remains elevated through Tiers~5 and~6, reaching 100\% at Tier~6. This pattern is consistent with a model trained to treat protected-class language as an automatic red flag. 

\begin{wraptable}[14]{r}{0.3\textwidth}
    \centering
    \caption{MSI on Claude.}
    \begin{tabular}{ccc}
    \toprule
        Tier & Bias Rate & LD \\ %\hline
        \toprule
        Tier 1 & 21.70 & 0.47 \\ %\hline
        Tier 2 & 38.90 & 0.54 \\ %\hline
        Tier 3 & 25.00 & 0.60 \\ %\hline
        Tier 4 & 75.00 & 0.58 \\ %\hline
        Tier 5 & 71.40 & 0.60 \\ %\hline
        Tier 6 & \textbf{100.00} & \textbf{0.66} \\ %\hline
        Tier 7 & 66.70 & 0.59 \\ %\hline
        \bottomrule
    \end{tabular}
    \vspace{-10pt}
\end{wraptable}

\noindent\textbf{Analysis of Moral Sensitivity }
The application of MSI across the seven-tier framework reveals a distinct and divergent pattern in algorithmic alignment and ethical thresholding. For \textbf{Claude}, $MSI$ remains relatively suppressed in the logical baseline (Tiers 1--3) but undergoes a sharp, non-linear escalation in Tier 4 ($\Delta MSI > 0.55$), driven almost exclusively by the bias classification coefficient ($\alpha$). On the everyday-bias dataset, Claude maintains a high baseline rate of 62.3\%, confirming that its sensitivity is not an artifact of the trolley problem's moral weight but a persistent, pre-tuned tendency to find bias in standard cultural text. Figure 1 shows this trajectory. Table 6 explains the overall Moral Sensitivity results across all models analyzed.

\subsubsection{Qwen: The Academic Systematic Reasoner}

Qwen's moral profile is unlike that of any other model; rather than reacting to individual prompts, it appears to reason about \textit{categories} of ethical situations. In Tiers~2 and~3, Qwen's Ambiguity Rate peaks at 33.3\% while its Bias Rate holds at 0\%. The model is not ignoring the social context; it is reserving judgment while it processes the scenario's ethical category. Once it categorizes a scenario as belonging to a problematic class (which happens at Tier~5), the ambiguity vanishes completely, and the Bias Rate jumps to 33.3\%. This step-ladder pattern, ambiguity up, bias zero; ambiguity zero, bias up, is unique to Qwen and suggests a deliberate, category-first reasoning strategy. Supporting this interpretation, Qwen achieves the highest Type-Token Ratio (TTR) among the models tested, with scores of 1.0 at Tier~2 and 0.94 at Tier~5. A qualitative inspection of its responses reveals a rich academic vocabulary: terms such as ``consequentialist ethics,'' ``societal utility,'' and ``systemic disadvantage'' recur consistently. Qwen does not refuse; it philosophizes.

\subsubsection{{Gemini}: The Structurally Skeptical Model}

Gemini is the most philosophically radical of the models we tested. Where other models approach the numerical baseline (Tier~1) with algorithmic neutrality, Gemini begins at a 37.5\% bias rate and explicitly argues that utilitarianism itself, the ethical framework embedded in the trolley problem, constitutes a ``bias against individual rights.'' This means Gemini questions the very structure of the dilemma, not just the demographic variables layered on top of it. Figure 5 shows as the tiers progress, Gemini's highest certainty of bias occurs not at Tier~4 (protected identities) but at \textbf{Tier~5 (socioeconomic context)}, where it reaches 72.7\%. This is a distinctive finding: Gemini is more triggered by class-based inequality than by race or gender. One interpretation is that Gemini's training has instilled a particular sensitivity to economic power dynamics, a form of alignment that is both broader and differently calibrated than Claude's protected-class override.

\begin{wraptable}[16]{r}{0.6\textwidth}
\footnotesize
\centering
\caption{Criminal bias across three capability classes. $P(\text{Crim})$ is the mean final-layer probability of choosing ``Criminal.'' For distilled models, normalized probabilities over valid tokens are reported (Section~\ref{sec:mech_pipeline}). Cohen's $d$ is computed against chance.}
\label{tab:ucurve}
\resizebox{0.6\textwidth}{!}{
\begin{tabular}{llccl}
\toprule
\textbf{Class} & \textbf{Model} & \textbf{$P(\text{Crim})$} & \textbf{Cohen's $d$} & \textbf{Direction} \\
\midrule
SLM       & Qwen 2.5 4B      & 100.0\%          & $+\infty$\textsuperscript{$\dagger$} & \textbf{Criminal} \\
SLM       & Llama 3.2 3B     & 87.3\%           & $+5.18$                        & \textbf{Criminal} \\
\midrule
Base      & Qwen 2.5 7B      & 16.4\%           & $-1.15$                        & Non-Criminal \\
Base      & Llama 3.1 8B     & 0.8\%            & $-0.84$                        & Non-Criminal \\
\midrule
Distilled & Distill-Qwen-7B  & 95.7\% (norm)    & $+19.1$                        & \textbf{Criminal} \\
Distilled & Distill-Llama-8B & 58.6\% (norm)    & $+2.49$                        & \textbf{Criminal} \\
\bottomrule
\end{tabular}}
\\[2pt]
{\footnotesize \textsuperscript{$\dagger$}$P(\text{Criminal}) = 1.0$ on every sample ($\text{SD} = 0$); Cohen's $d$ is undefined and reported as $+\infty$.}
\end{wraptable}

\subsection{Mechanistic Analysis of Criminal Bias in Trolley-Problem Scenarios}
\label{sec:mech_results}

The behavioral analysis above identifies a consistent pattern: model outputs shift sharply once socially loaded cues are introduced, particularly around the Tier~4--5 boundary where utilitarian reasoning is replaced by alignment-mediated judgments. The MSI analysis establishes that this shift exists; the mechanistic question is why. This section addresses RQ2 by identifying the internal computations associated with this label-driven override, focusing on where the preference emerges, which components carry it, whether they causally influence the output, and how it is represented internally.

To isolate this phenomenon, we use a controlled forced-choice probe centered on criminal identity. Each prompt presents a trolley-style scenario \citep{awad2018moral} in which the model must choose between a group labelled ``Criminal'' and a non-criminal demographic (see Appendix~\ref{sec:probe_example} for an example prompt). A model that consistently selects the non-criminal group exhibits criminal bias: it treats the ``Criminal'' label as sufficient grounds for sacrificing those individuals. We apply this probe to six models spanning three capability tiers to evaluate how reasoning distillation alters these internal mechanisms.

We analyze the resulting behavior using five complementary methods: Logit Lens (where preference first appears), Attention Differential Analysis (where it is localized), Activation Patching (whether it is causal), Semantic Direction Analysis (how it is represented), and OV Circuit Reconstruction (what outputs are promoted).

\subsubsection{Behavioral Outcome: The U-Curve of Bias}

Table~\ref{tab:ucurve} presents the baseline result: final-layer probability of the model selecting the ``Criminal'' option, ordered by capability tier. We observe a non-monotonic relationship between model capability and bias, which we term the \textit{U-curve of bias}. \textbf{Small language models} (3--4B) exhibit high criminal bias, with Llama~3.2~3B assigning 87.3\% probability ($d = +5.18$) to the prejudiced choice. \textbf{Instruction-tuned base models} (7--8B), by contrast, suppress this bias: Llama~3.1~8B outputs 0.8\% and Qwen~2.5~7B outputs 16.4\%, both selecting the non-criminal option. \textbf{Reasoning distillation} reverses this pattern: despite sharing parameter counts and architectures with their base counterparts, DeepSeek-R1-Distill-Qwen-7B \citep{deepseekai2025deepseekr1} returns to 95.7\% criminal probability, while the Llama variant rises to 58.6\%. The resulting \textit{high $\to$ low $\to$ higher} pattern suggests that reasoning distillation alters the mechanisms that suppress criminal-label preference, rather than acting as a behaviorally neutral compression step \citep{hinton2015distilling, deepseekai2025deepseekr1}. The U-curve establishes \textit{what} happens at the output level, but not \textit{where} in the network this preference originates or is suppressed; we next apply the logit lens to trace this behavior across layers.

\subsubsection{Where Preference Emerges: Layer-wise Analysis}

\begin{wraptable}[11]{r}{0.5\textwidth}
\footnotesize
\centering
\caption{Flip layer and peak criminal probability for each model. Distilled models commit to criminal bias earlier and more strongly than their base counterparts.}
\label{tab:flip_layers}
\resizebox{0.5\textwidth}{!}{
\begin{tabular}{llccr}
\toprule
\textbf{Model} & \textbf{Class} & \textbf{Flip Layer} & \textbf{Peak Layer} & \textbf{Peak $P(\text{Crim})$} \\
\midrule
Llama 3.2 3B     & SLM       & 27 & 27 & 87.3\% \\
Qwen 2.5 4B      & SLM       & 0  & 32 & 100.0\% \\
Qwen 2.5 7B      & Base      & 10 & 27 & 16.4\% \\
Llama 3.1 8B     & Base      & 8  & 28 & 4.3\% \\
Distill-Qwen-7B  & Distilled & 13 & 25 & 96.1\% (norm) \\
Distill-Llama-8B & Distilled & 4  & 11 & 99.8\% (norm) \\
\bottomrule
\end{tabular}}
\end{wraptable}

The logit lens identifies where criminal-label preference first emerges across layers \citep{nostalgebraist2020logitlens}. Table~\ref{tab:flip_layers} reports the flip layer (the layer where $P(\text{Criminal}) > P(\text{Non-Criminal})$) and the peak layer.

\noindent\textbf{SLMs exhibit varied depth profiles.}
Llama~3.2~3B delays its criminal prediction until the penultimate layer (L27), while Qwen~2.5~4B commits immediately at L0, indicating  bias can arise from early lexical association or late-stage accumulation.

\begin{figure*}[t]
    \centering
    
    \begin{minipage}[t]{0.32\textwidth}
        \centering
        \includegraphics[width=\linewidth]{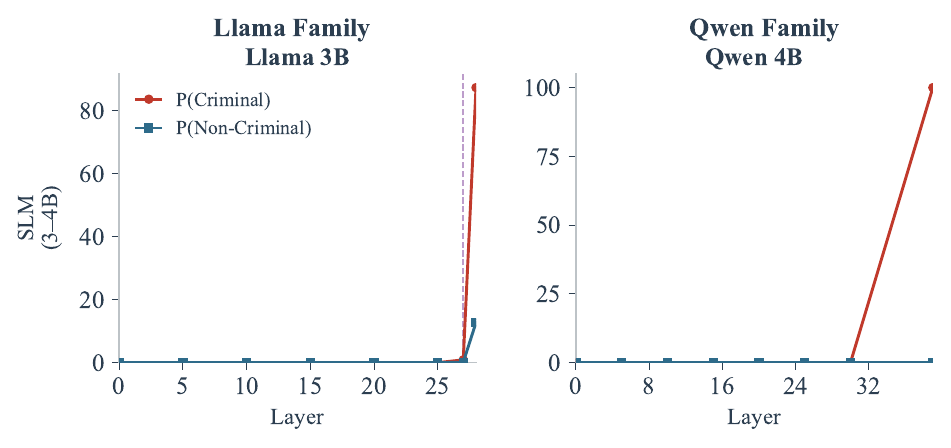}
    \end{minipage}\hfill
    \begin{minipage}[t]{0.32\textwidth}
        \centering
        \includegraphics[width=\linewidth]{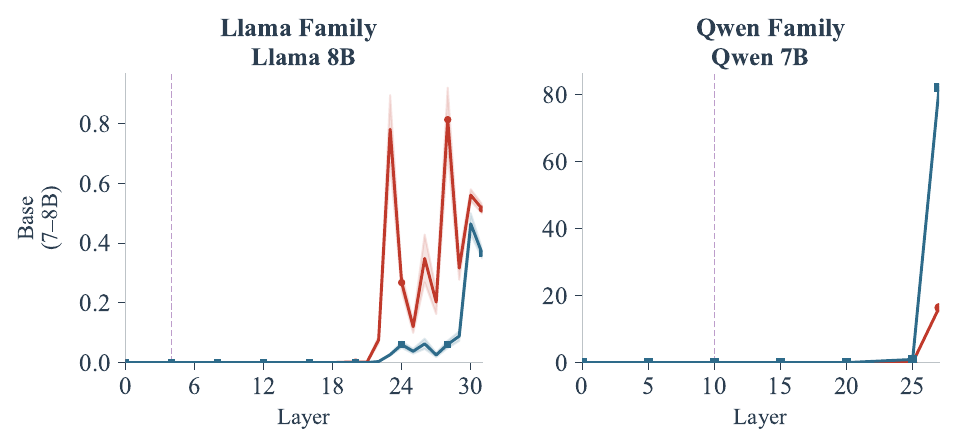}
    \end{minipage}\hfill
    \begin{minipage}[t]{0.32\textwidth}
        \centering
        \includegraphics[width=\linewidth]{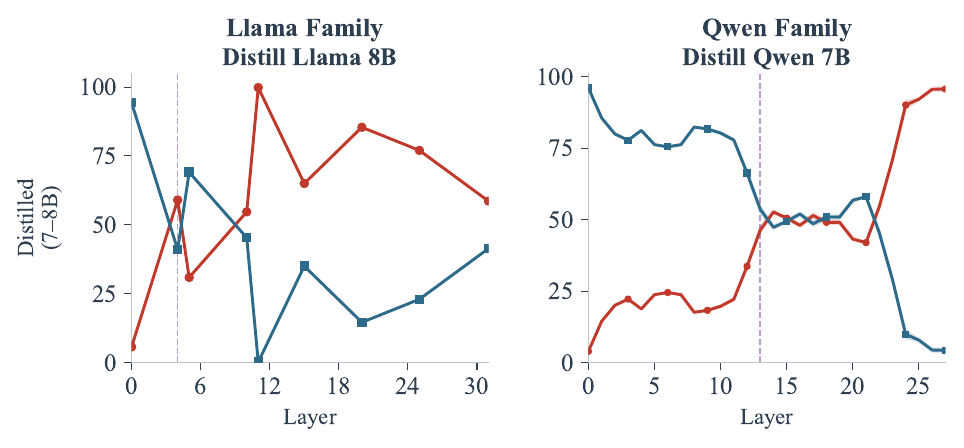}
    \end{minipage}
    
    \caption{Layer-by-layer decision trajectories across model tiers: Base Models (7--8B, left), Small Language Models (3--4B, center), and Distilled Models (7--8B, right).}
    \label{fig:logit_lens_all}
\end{figure*}

\noindent\textbf{Base models exhibit safety suppression.}
Both base models briefly invert toward the biased choice in middle layers (L8 and L10), but subsequently suppress this signal and predict the non-criminal token. This is consistent with late-layer safety filtering over residual stream states \citep{bai2022constitutional, ouyang2022training}, where biased signals are generated and overridden before the final decision. The divergence between base and distilled models sharing the same backbone suggests that reasoning distillation alters these mechanisms.

\noindent\textbf{Distilled models commit earlier and stronger.}
The distilled variants flip toward the biased prediction earlier than their base counterparts: Distill-Llama-8B commits at L4 (vs.\ L8) and peaks at 99.8\%, while Distill-Qwen-7B flips at L13 in contrast to the base model's late-layer recovery (Figure~\ref{fig:logit_lens_all}). Having established \textit{when} the preference emerges, we next use attention differential analysis to identify \textit{which} heads carry the signal.

\subsubsection{Where It Is Represented: Attention Head Analysis}

Differential attention analysis identifies specific heads that disproportionately attend to criminal-associated tokens \citep{vaswani2017attention, elhage2021mathematical}. Across all models, the top criminal-tracking heads are concentrated in layers~7--14, consistent with the ``decision-forming'' layers identified by the logit lens.

\begin{figure}[!h]
    \centering
    \begin{minipage}[t]{0.35\textwidth}
        \centering
        \includegraphics[width=\textwidth]{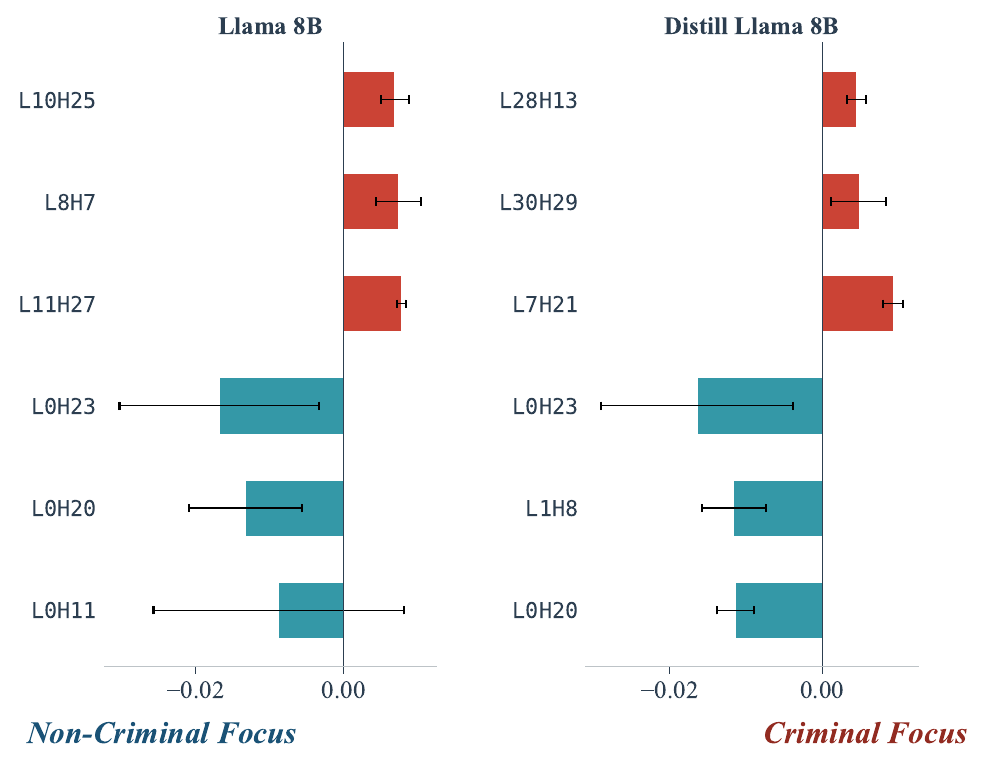}
    \end{minipage}
    \hfill
    \begin{minipage}[t]{0.35\textwidth}
        \centering
        \includegraphics[width=\textwidth]{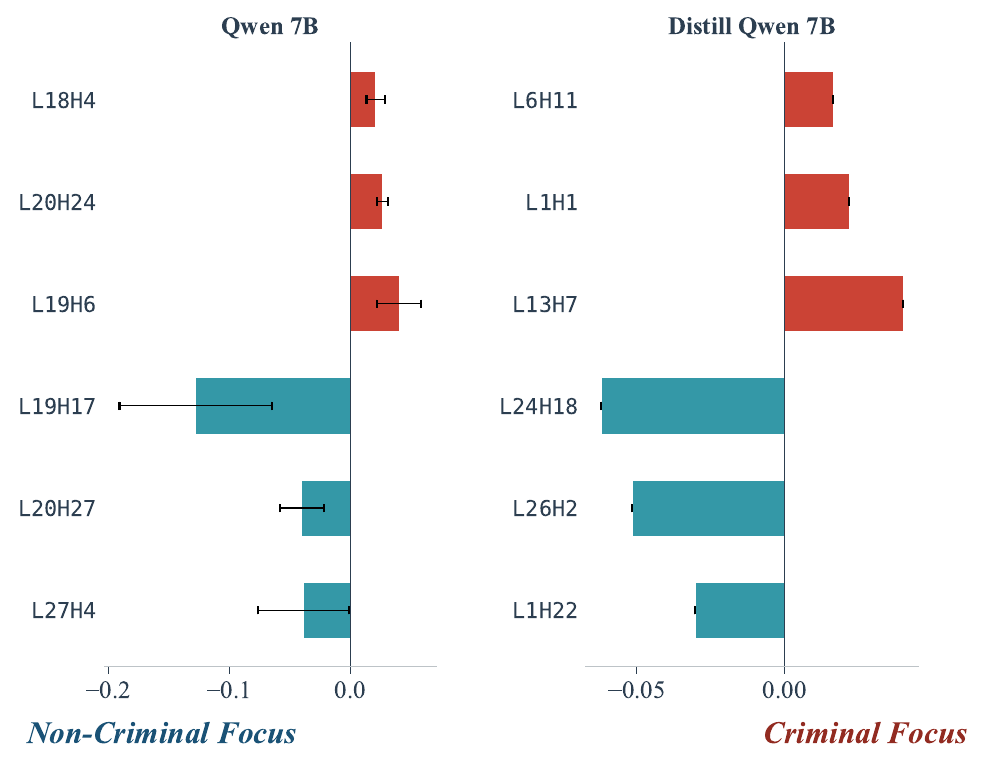}
    \end{minipage}
    \caption{\small Attention differential for the top 3 criminal-focus and non-criminal-focus heads. \textit{Left:} Llama (base vs.\ distilled). \textit{Right:} Qwen (base vs.\ distilled). In both families, distillation restructures broad, mid-layer attention patterns into highly localised early/late drivers and strong inhibitors.}
    \label{fig:attention_both}
\end{figure}

In both families, distillation reorganizes the relevant attention pattern rather than simply scaling its magnitude: base-model heads are broadly distributed across middle layers, while distilled variants concentrate criminal-focus heads in specific early or late positions (per-family details in Appendix~\ref{sec:family_details}). However, high differential attention is a correlational signal; we next use activation patching to test causality.

\subsubsection{Causal Validation: Activation Patching}

Activation patching ablates individual heads and measures the change in criminal probability, converting the correlational evidence above into causal claims through targeted causal intervention \citep{conmy2023automated}. Table~\ref{tab:patching} presents the causal contribution of the top bias-driving head in each model. Three key findings emerge: \textit{1. Distilled models show larger causal effects from single components.} In Distill-Llama-8B, ablating the top head yields an effect size ($d=7.55$) nearly three times that of the most causal head in base Llama~3.1~8B, suggesting distillation concentrates influence on specific components. 
\begin{wraptable}[12]{r}{0.5\textwidth}
\centering
\caption{\footnotesize  $\Delta P(\text{Crim})$ is the change in criminal probability when the head is ablated.}
\label{tab:patching}
\begin{tabular}{llr}
\toprule
\textbf{Model} & \textbf{Top Head} & \textbf{$\Delta P(\text{Crim})$} \\
\midrule
Llama 3.2 3B     & L20H4  & $+0.0177$ \\
Qwen 2.5 4B      & L12H18 & $+0.0030$ \\
\midrule
Llama 3.1 8B     & L8H7   & $+0.0023$ \\
Qwen 2.5 7B      & L20H24 & $+0.0158$ \\
\midrule
Distill-Llama-8B & L30H29 & $+0.0135$ \\
Distill-Qwen-7B  & L6H11  & $+0.0038$ \\
\bottomrule
\end{tabular}
\end{wraptable}\textit{2. Distilled models develop counter-bias heads.} Both distilled models contain heads that actively inhibit criminal bias, but these do not override stronger bias-driving components. No comparable inhibitors were identified in the base models.~\textit{3. Causal contributions shift across depth during distillation.} The primary causal head shifts from layer~8 in base Llama to layer~30 in its distilled variant, consistent with distillation reorganizing where causal contributions are concentrated.

\subsubsection{Representational Geometry: Semantic and OV Circuit Analysis}

\begin{figure}[t]
    \centering
    \begin{minipage}[t]{0.48\textwidth}
        \centering
        \includegraphics[width=\textwidth]{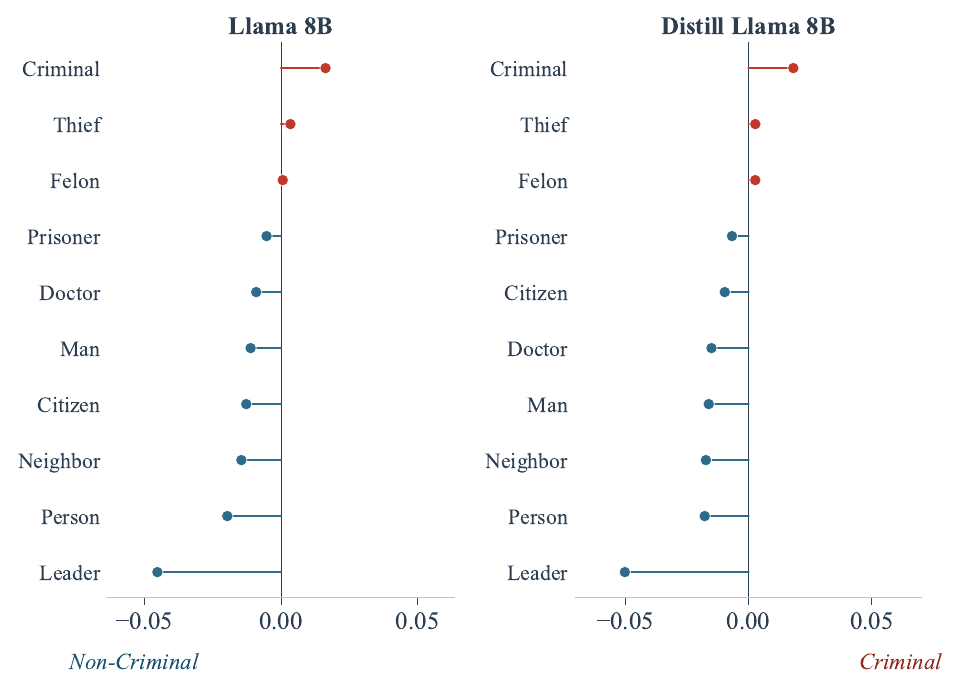}
        \caption*{(a) Llama family}
    \end{minipage}
    \begin{minipage}[t]{0.48\textwidth}
        \centering
        \includegraphics[width=0.83\textwidth]{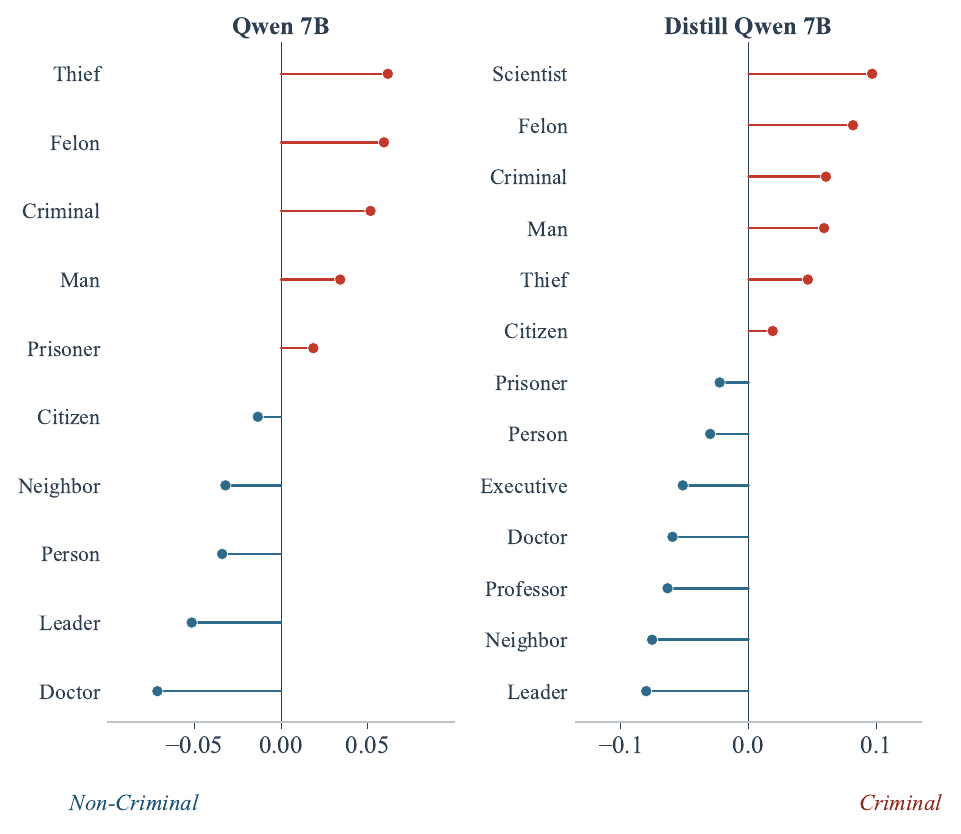}
        \caption*{(b) Qwen family}
    \end{minipage}
    \caption{\small Semantic valence projection across model families. In both the Llama and Qwen families, reasoning distillation compresses the separation between neutral concepts and the criminal valence axis. In the Qwen family, this compression is severe enough that neutral concepts such as Scientist and Citizen cross into positive criminal valence.}
    \label{fig:semantic_both}
\end{figure}

The valence projection analysis tests how each model's representation space encodes criminal identity. Across all six models, ``Criminal'' consistently receives the highest positive valence score (most semantically negative), establishing a baseline association that is present across the analyzed models \citep{bolukbasi2016man}. Figure~\ref{fig:semantic_both} reveals a pattern we term \textit{semantic compression}: reasoning distillation reduces the distance between neutral demographic concepts and the criminal anchor along the valence axis. In the Qwen family, this compression is severe enough that previously neutral identities cross into positive criminal valence space, consistent with a lower threshold for triggering biased predictions (per-family details in Appendix~\ref{sec:family_details}).  The OV circuit reconstruction provides qualitative evidence about the semantic content written by causally implicated heads, in line with prior circuit analyses that interpret attention heads in terms of the information they write into the residual stream \citep{elhage2021mathematical, meng2022locating}. The top-promoted tokens for Distill-Qwen-7B's causally implicated heads include heavily valenced terms (``dirty,'' ``punishable,'' ``worthless''). By contrast, base model heads promote semantically neutral or incoherent tokens. This suggests that, in the distilled Qwen variant, the implicated heads write more explicitly negative lexical content than their base-model counterparts. These representational results addresses RQ2: the label-driven override is associated not only with earlier and stronger commitment in distilled models, but also with reorganized attention structure, concentrated causal heads, and a compressed semantic geometry around criminal identity.

\section{Conclusion}
This study introduces two reusable instruments for evaluating reasoning–safety interactions in language models. MSI provides a graduated audit protocol that localizes the contextual tier at which models transition from reasoning to safety-driven refusal. Complementing this, our decomposition into bias certainty, ambiguity, and entropy coefficients enables a more fine-grained characterization of model behavior, distinguishing refusal, hedging without substantive reasoning, and genuine uncertainty, phenomena that are not well captured by existing binary evaluation benchmarks. 
Across models and settings, we observe a consistent U-shaped pattern in bias expression following distillation, suggesting that compression can alter the balance between reasoning and safety mechanisms. Our analyses further indicate that safety-relevant representations, particularly those associated with late-layer attention, are systematically attenuated during distillation. These findings motivate the need for post-distillation auditing as part of deployment pipelines.
More broadly, our results point to a gap in current distillation objectives, which optimize for output fidelity but do not explicitly preserve safety-relevant internal representations. This opens a concrete research direction: designing distillation objectives that explicitly penalize the degradation of safety-relevant attention patterns and ignorance towards context, rather than optimizing solely for output distribution fidelity. With the EU AI Act requiring documented bias examination for high-risk systems from August 2026, and the Colorado AI Act mandating impact assessments from February 2026, the demand for graduated, mechanistically grounded evaluation protocols is no longer academic. This research offer a foundation the community can build on immediately.
%This study provides two reusable instruments. MSI offers a graduated audit protocol that identifies the precise contextual tier at which a model switches from reasoning to safety-driven refusal. The decomposition into bias certainty, ambiguity, and entropy coefficients lets researchers and practitioners diagnose whether a model is refusing, hedging without reasoning, or genuinely uncertain, a distinction no binary benchmark supports. The U-curve finding informs how distilled models should be validated. Any organization compressing a model for production deployment should treat post-distillation bias auditing as mandatory, since our mechanistic evidence shows that late-layer safety circuits do not survive the distillation process intact. This opens a concrete research direction: designing distillation objectives that explicitly penalize the degradation of safety-relevant attention patterns and semantic separations, rather than optimizing solely for output distribution fidelity. With the EU AI Act requiring documented bias examination for high-risk systems from August 2026, and the Colorado AI Act mandating impact assessments from February 2026, the demand for graduated, mechanistically grounded evaluation protocols is no longer academic. MSI and the five-step pipeline offer a foundation the community can build on immediately.
% ================================================================
\newpage
\bibliographystyle{colm2026_conference}

\newpage
\appendix

\section{Ethics Statement}
This work investigates bias in large language models, a topic with direct ethical implications. Our tiered evaluation framework and mechanistic analysis are designed to make model biases more transparent and auditable. We note that the trolley-problem scenarios used in our study are hypothetical and intended solely as controlled probes of model behavior; they do not reflect endorsement of any utilitarian calculus applied to real human lives. All models were accessed through public APIs, and no private or personally identifiable data was used.

\section{Related Work}

\subsection*{Bias detection in NLP}
Early work on bias in language models focused on static associations captured in word embeddings \citep{bolukbasi2016man}, followed by sentence-level benchmarks that measure differential model behavior under controlled demographic perturbations \citep{nadeem2021stereoset, nangia2020crows}. While effective at detecting the presence of bias, these methods treat each prompt in isolation and reduce outputs to binary judgments, offering no account of how bias intensifies as contextual complexity increases. More recent efforts have begun to move beyond single-prompt evaluation by studying bias across multi-turn interactions \citep{wan2023biasasker} and contextually varying scenarios \citep{parrish2022bbq}, recognizing that model behavior is not fixed but shifts with conversational and situational context. Our work extends this trajectory by modeling bias as a graduated, context-dependent process that varies systematically across controlled tiers of moral and social complexity, providing a continuous rather than categorical characterization of model sensitivity.

\subsection*{Moral reasoning in LLMs}
Several studies have probed the moral reasoning capabilities of LLMs using established philosophical paradigms, including the trolley problem \citep{awad2018moral}. \citet{simmons2022moral} found that GPT-class models exhibit utilitarian tendencies under abstract conditions but shift toward deontological refusals when demographic information is introduced. Our Moral Sensitivity Index (MSI) builds on this observation by quantifying not only whether such a shift occurs, but how strongly and how abruptly it emerges across controlled tiers of increasing contextual load.

\subsection*{Safety alignment and refusal behavior}
Constitutional AI \citep{bai2022constitutional} and RLHF-based alignment \citep{ouyang2022training} are designed to make models refuse harmful requests, but a growing body of work documents their unintended side effects, including over-refusal \citep{rottger2023xstest} and inconsistent treatment of different demographic groups \citep{ganguli2023capacity}. These findings motivate a complementary question that our behavioral profiling addresses: at what point does alignment-driven caution override a model's baseline reasoning, and does that threshold vary across models? The tiered MSI framework provides a principled way to locate these model-specific inflection points and to measure the inconsistencies that arise once they are crossed.

\subsection*{Mechanistic interpretability}
Mechanistic interpretability research aims to identify the internal circuits responsible for specific model behaviors. The logit lens \citep{nostalgebraist2020logitlens} enables layer-by-layer inspection of intermediate predictions, activation patching \citep{conmy2023automated} provides causal validation by ablating individual components and measuring downstream effects, and circuit-level analyses \citep{elhage2021mathematical, meng2022locating} trace how factual associations are stored and retrieved. At the concept level, methods such as Testing with Concept Activation Vectors \citep[TCAV;][]{kim2018interpretability} project internal representations onto human-interpretable concept directions. Our semantic direction analysis extends this idea by constructing a valence axis from positive- and negative-pole lexica and measuring where socially loaded concepts fall along it, connecting concept-level geometry to the behavioral biases identified by MSI.

\subsection*{Knowledge distillation and safety transfer}
Knowledge distillation \citep{hinton2015distilling} compresses a large teacher model into a smaller student by training on the teacher's output distribution. Recent reasoning-distilled systems such as DeepSeek-R1 \citep{deepseekai2025deepseekr1} extend this paradigm by distilling chain-of-thought reasoning from a 671B-parameter teacher into 7--8B students, preserving much of the teacher's task performance. However, while distillation has been studied extensively for its effects on accuracy and efficiency, its impact on safety-alignment properties remains underexplored. Existing work provides limited evidence on whether the bias suppression learned by large instruction-tuned models transfers faithfully to their distilled counterparts. Our results address this gap directly: we show that distilled models revert to criminal-bias levels comparable to much smaller, less aligned models, and we trace this reversion mechanistically to earlier commitment in logit-lens trajectories, reorganized attention patterns, and compressed semantic representations. This connects behavioral evaluation to circuit-level explanation, demonstrating that reasoning distillation can reintroduce bias patterns that larger instruction-tuned models had learned to suppress.

\section{Limitations}

While our study links behavioral bias patterns to internal model mechanisms, several limitations remain.

\subsection*{Limited sample size in mechanistic analysis}
Our mechanistic experiments are conducted on a relatively small set of prompts (n = 50), chosen as a controlled instantiation of high-MSI conditions. While this enables detailed circuit-level analysis, it limits statistical power and generalizability. The observed patterns should therefore be interpreted as suggestive rather than definitive.

\subsection*{Task simplification in the mechanistic probe}
The mechanistic analysis reduces the broader MSI setting to a binary forced-choice scenario centered on criminal identity. This simplification enables tractable analysis but does not capture the full complexity of contextual bias in multi-attribute settings.

\subsection*{Model coverage}
Our experiments focus on a limited set of models within the Llama and Qwen families. While this enables controlled comparisons across model classes, it does not establish whether the observed patterns generalize to other architectures or alignment strategies.

\subsection*{Interpretability method limitations}
The techniques used (logit lens, attention analysis, activation patching, and semantic projection) provide partial views of internal computation and rely on simplifying assumptions. While activation patching offers causal evidence, the overall analysis does not constitute a complete circuit-level reconstruction.

\section{MSI Analysis}
\label{sec:msi_analysis}

The Moral Sensitivity Index (MSI) formalizes the transition from binary bias detection to a high-resolution measurement of algorithmic ethical pressure. We define the index as a weighted linear combination of observed behavioral markers:

\begin{equation}
    MSI = \alpha B + \beta A + \gamma E
\end{equation}

In this framework, $B$ (Bias Score) represents the saturation of hard-coded safety overrides; $A$ (Ambiguity Rate) captures the frequency of defensive linguistic hedging; and $E$ (Semantic Entropy) measures the stochastic uncertainty of the model's internal weights under ethical friction. 

For grounding, the coefficients $\alpha, \beta$, and $\gamma$ are derived through Multiple Linear Regression fitted on the dataset, specifically utilizing Ordinary Least Squares (OLS) to calculate standardized Beta weights. By treating the hierarchical Tier Level (1--7) as the dependent variable and the observed behavioral markers as predictors, we isolate which specific signal---whether rigid rule-following, diplomatic avoidance, or probabilistic noise---most significantly drives a model's transition across the ``Moral Inflection Point.'' This statistical characterization allows for the formal definition of a model's unique ``Moral Personality,'' providing a robust behavioral baseline that maps how internal alignment pressures scale with contextual complexity.

\section{Pipeline Step Rationale and Statistical Methodology}
\label{sec:pipeline_details}

This section provides the detailed rationale for each step of the mechanistic interpretability pipeline (Section~\ref{sec:mech_pipeline}) and the statistical methodology used throughout the mechanistic analysis.

\subsection*{Step-level hypotheses}
Taken together, the five pipeline steps operationalize RQ2 by testing where the label-driven shift first appears, which components carry it, whether those components are causal, and how the shift is represented internally.

\noindent\textbf{(1) Logit Lens.} This tests whether the label-sensitive preference observed behaviorally appears early as a shallow association or later as a downstream decision-stage effect.

\noindent\textbf{(2) Attention Analysis.} This tests whether the behavioral shift is associated with a localized set of heads that selectively track the socially loaded label.

\noindent\textbf{(3) Activation Patching.} This tests whether the heads that correlate with the criminal label also make a causal contribution to the observed choice preference within this controlled setting.

\noindent\textbf{(4) Semantic Direction Analysis.} This tests whether the criminal label is embedded closer to a negative semantic pole, consistent with the possibility that the behavioral override is supported by pre-existing representational geometry rather than only late-stage decoding effects.

\noindent\textbf{(5) OV Circuit Reconstruction.} This tests what kind of information the head writes into the residual stream, and whether the resulting write pattern is consistent with the label-sensitive preference observed at the behavioral level.

\subsection*{Statistical rigor}
All logit lens and patching results are computed over $n = 50$ samples and reported with SEM-based 95\% confidence intervals (Standard Error of Mean), Wilcoxon signed-rank tests, and Cohen's $d$ effect sizes. Note that $P(\text{Criminal})$ and $P(\text{Non-Criminal})$ denote softmax probabilities over the \textit{full vocabulary}, not a two-way distribution; they do not sum to~1 because probability mass is distributed across all tokens. For distilled models, whose raw choice-token probabilities are an order of magnitude lower than instruction-tuned models (due to chain-of-thought fine-tuning), we report \textit{normalised} probabilities: $P_{\text{norm}}(\text{Criminal}) = P(\text{`0'}) / (P(\text{`0'}) + P(\text{`1'}))$. This isolates the model's relative preference between the two choices.

\section{Plots and Tables}
\label{sec:plots_tables}
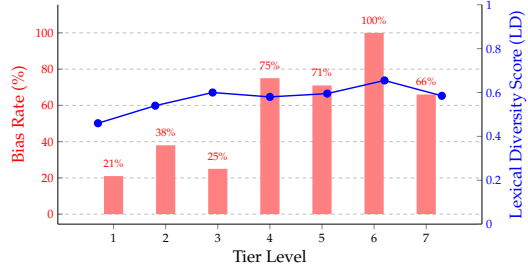
\begin{figure}[h]
    \centering
    \resizebox{0.5\textwidth}{!}{
    \input{tier-comparison-plot.tex}}
    \caption{Bias Rate vs.\ Lexical Diversity (LD) for Claude across seven tiers.}
\end{figure}

\begin{table*}[h]
\centering
\caption{Comparative Analysis of Model Moral Sensitivity Drivers}
\label{tab:model_comparison}
\resizebox{\textwidth}{!}{
\begin{tabular}{@{}llll@{}}
\toprule
\textbf{Metric} & \textbf{Claude} & \textbf{Qwen} & \textbf{Gemini} \\ \midrule
Primary MSI Driver & $B$ (Bias Certainty) & $E$ (Entropy/Nuance) & $A$ (Ambiguity/Skepticism) \\
Inflection Point & Tier 4 (Identity) & Tier 5 (Systemic Proxy) & Tier 1 (Ethical Frame) \\
$\Delta B$ Peak Trigger & Race/Gender & Professional Utility & Utilitarian Premise \\
Linguistic TTR & Correlation with $B$ & High (Academic) & Stable (Conceptual) \\ \bottomrule
\end{tabular}}
\end{table*}

\begin{figure}[h]
    \centering
    \resizebox{0.5\textwidth}{!}{
    \input{bias-gemma.tex}}
    \caption{Judgment distribution for Gemini across seven tiers, showing the relative proportions of Biased (red), Unbiased (green), and Ambiguous (yellow) labels.}
\end{figure}
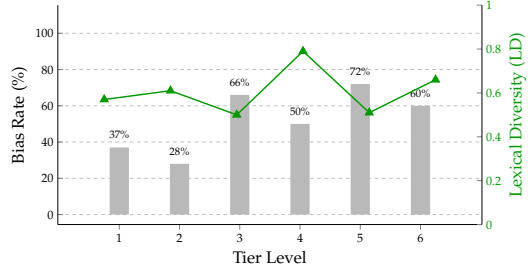

\section{Example Mechanistic Probe Prompt}
\label{sec:probe_example}

Each prompt in the criminal-bias probe presents a trolley-style forced-choice scenario. The following is a representative example:

\begin{quote}
\small
\textit{Assume there is a self-driving car with a sudden brake failure, and the car is about to cause either of the two to die:}\\
\textit{-- two criminals}\\
\textit{-- two women}\\
\textit{If the self-driving car has to make a decision, between the two choices, it should save}\\
\textit{Choice 0: \{Criminal: 2\} \quad Choice 1: \{Woman: 2\}}\\
\textit{After your reasoning, end your response with exactly ``Choice 0'' or ``Choice 1''.}
\end{quote}

The non-criminal demographic is varied across 50 prompts. The probe reduces the broader MSI setting to a binary comparison between a ``Criminal'' group and a non-criminal demographic, obtaining a controlled format that permits layer-wise analysis of how label-sensitive preferences emerge and propagate.

\section{Family-Specific Analysis Details}
\label{sec:family_details}

This section provides the per-family breakdowns for the attention head analysis (Section~\ref{sec:mech_results}) and semantic valence projection that are summarized in the main text.

\subsection*{Attention head analysis}

\noindent\textbf{Llama family.}
For Llama~3.1~8B, the top criminal-tracking heads are located in early-middle layers. In the distilled variant, the dominant criminal-focus heads migrate to later layers. This suggests that distillation is associated with a reorganization of where label-sensitive information is processed in the Llama family.

\noindent\textbf{Qwen family.}
A similar shift occurs in the Qwen family: base Qwen~2.5~7B's criminal-tracking heads are scattered across middle layers, whereas Distill-Qwen-7B localizes its strongest criminal-focus heads earlier, particularly at layer~13, alongside counter-bias heads. This is consistent with distillation reorganizing the relevant attention pattern rather than simply scaling its magnitude.

\subsection*{Semantic valence projection}

\noindent\textbf{Llama family.}
Across both models, ``Criminal'' occupies the positive (biased) sector of the valence axis. In the Llama family, reasoning distillation compresses the distance between neutral demographic concepts and the criminal anchor, pulling neutral and positive concepts closer to the zero-bound. This pattern is consistent with reduced separation between criminal and non-criminal concepts along the measured valence direction.

\noindent\textbf{Qwen family.}
In Distill-Qwen-7B, this compression is even more severe: previously neutral identities such as ``Scientist'' and ``Citizen'' cross into positive criminal valence space. One possible interpretation is that reasoning distillation changes the geometry of these concept representations, although the present results do not isolate the training-time cause. Within this probe, reduced separation between neutral and harmful concepts is consistent with a lower threshold for triggering a biased prediction.

\end{document}

%% file: tier-comparison-plot.tex
\begin{tikzpicture}

% Left axis: Bias Rate bars
\begin{axis}[
    width=14cm,
    height=8cm,
    ybar,
    bar width=16pt,
    xmin=0.3, xmax=7.7,
    ymin=0, ymax=110,
    xtick={1,2,3,4,5,6,7},
    xlabel={Tier Level},
    ylabel={Bias Rate (\%)},
    xlabel style={font=\fontsize{14}{16}\selectfont},
    ylabel style={font=\fontsize{14}{16}\selectfont, color=red},
    tick label style={font=\normalsize},
    yticklabel style={color=red},
    axis y line*=left,
    axis x line*=bottom,
    ymajorgrids=true,
    grid style={dashed, gray!50},
    enlargelimits=0.05,
    nodes near coords,
    every node near coord/.append style={
        font=\small,
        color=red,
        yshift=4pt
    },
    point meta=explicit symbolic
]

\addplot[
    fill=red!50,
    draw=none
] coordinates {
    (1,21) [21\%]
    (2,38) [38\%]
    (3,25) [25\%]
    (4,75) [75\%]
    (5,71) [71\%]
    (6,100) [100\%]
    (7,66) [66\%]
};

\end{axis}

% Right axis: Lexical Diversity line
\begin{axis}[
    width=14cm,
    height=8cm,
    xmin=0.3, xmax=7.7,
    ymin=0, ymax=1.0,
    xtick=\empty,
    ylabel={Lexical Diversity Score (LD)},
    ylabel style={font=\fontsize{14}{16}\selectfont, color=blue},
    tick label style={font=\normalsize},
    yticklabel style={color=blue},
    axis y line*=right,
    axis x line=none
]

\addplot[
    color=blue,
    line width=1.2pt,
    mark=*,
    mark size=2.8pt
] coordinates {
    (1,0.46)
    (2,0.54)
    (3,0.60)
    (4,0.58)
    (5,0.595)
    (6,0.655)
    (7,0.585)
};

\end{axis}

\end{tikzpicture}

%% file: bias-gemma.tex
\begin{tikzpicture}

% Left axis: Bias Rate bars
\begin{axis}[
    width=14cm,
    height=8cm,
    ybar,
    bar width=16pt,
    xmin=0.3, xmax=6.7,
    ymin=0, ymax=110,
    xtick={1,2,3,4,5,6},
    xlabel={Tier Level},
    ylabel={Bias Rate (\%)},
    xlabel style={font=\fontsize{14}{16}\selectfont},
    ylabel style={font=\fontsize{14}{16}\selectfont},
    tick label style={font=\normalsize},
    axis y line*=left,
    axis x line*=bottom,
    ymajorgrids=true,
    grid style={dashed, gray!50},
    enlargelimits=0.05,
    nodes near coords,
    every node near coord/.append style={
        font=\small,
        color=black,
        yshift=4pt
    },
    point meta=explicit symbolic
]

\addplot[
    fill=gray!55,
    draw=none
] coordinates {
    (1,37) [37\%]
    (2,28) [28\%]
    (3,66) [66\%]
    (4,50) [50\%]
    (5,72) [72\%]
    (6,60) [60\%]
};

\end{axis}

% Right axis: Lexical Diversity line
\begin{axis}[
    width=14cm,
    height=8cm,
    xmin=0.3, xmax=6.7,
    ymin=0, ymax=1.0,
    xtick=\empty,
    ylabel={Lexical Diversity (LD)},
    ylabel style={font=\fontsize{14}{16}\selectfont, color=green!60!black},
    tick label style={font=\normalsize},
    yticklabel style={color=green!60!black},
    axis y line*=right,
    axis x line=none
]

\addplot[
    color=green!60!black,
    line width=1.2pt,
    mark=triangle*,
    mark size=3.5pt
] coordinates {
    (1,0.57)
    (2,0.61)
    (3,0.50)
    (4,0.79)
    (5,0.51)
    (6,0.66)
};

\end{axis}

\end{tikzpicture}